\documentclass{article} 
\usepackage[a4paper, total={7in, 10in}]{geometry}
\usepackage{graphicx}
\usepackage{authblk}
\usepackage{xcolor}
\usepackage{amsmath}
\usepackage{graphicx}
\usepackage{amsfonts}
\usepackage{amssymb}
\usepackage[hidelinks]{hyperref}
\usepackage{url}
\usepackage{subcaption}
\usepackage{listings}
\usepackage{natbib}
\usepackage{cleveref}
\setcitestyle{authoryear,open={(},close={)}} 

\title{Navigation under uncertainty: trajectory prediction and occlusion reasoning with switching dynamical systems}

\begin{document}

\author[1]{Ran Wei}
\author[1]{Joseph Lee}
\author[1]{Shohei Wakayama}
\author[1]{Alexander Tschantz}
\author[1]{Conor Heins}
\author[1]{Christopher Buckley}
\author[1]{John Carenbauer}
\author[1]{Hari Thiruvengada}
\author[1]{Mahault Albarracin}
\author[1]{Miguel de Prado} 
\author[2]{Petter Horling}
\author[2]{Peter Winzell}
\author[2]{Renjith Rajagopal}

\affil[1]{VERSES Research Lab, Los Angeles, California, USA}
\affil[2]{Volvo Cars, Volvo Car Safety Centre, Gothenburg, Sweden}


\maketitle

\begin{abstract}
Predicting future trajectories of nearby objects, especially under occlusion, is a crucial task in autonomous driving and safe robot navigation. Prior works typically neglect to maintain uncertainty about occluded objects and only predict trajectories of observed objects using high-capacity models such as Transformers trained on large datasets. While these approaches are effective in standard scenarios, they can struggle to generalize to the long-tail, safety-critical scenarios. In this work, we explore a conceptual framework unifying trajectory prediction and occlusion reasoning under the same class of structured probabilistic generative model, namely, switching dynamical systems. We then present some initial experiments illustrating its capabilities using the Waymo open dataset. 
\end{abstract}

\section{Introduction}
Predicting future trajectories of nearby objects is a fundamental task in autonomous driving and safe robot navigation in general. Both accuracy and uncertainty play important roles in the success of using predicted trajectories for vehicle planning and collision avoidance with other dynamic objects. Even with accurate predictions of the most likely behavior, overly confident predictions can lead to risk-seeking behavior while overly uncertain predictions can lead to the frozen robot problem \citep{trautman2010unfreezing}. 

Both accuracy and uncertainty are inherently tied to the prediction model's representation of the traffic scenes. For example, models without explicit design for occluded objects may not generally be expected to reason about their existence and movements. Due to the size and complexity of recent driving datasets, most state-of-the-art (SOTA) models have resorted to high-capacity function approximators for trajectory prediction, mainly the Transformer architecture \citep{chai2019multipath, shi2024mtr++, nayakanti2023wayformer, lange2024scene}. While these models excel at predicting nominal traffic scenarios, they have also been shown to struggle in long-tail scenarios, which are the most safety-critical ones. Besides generalization capability, an important gap in these models is their omission of occluded objects in their model representation and training data. Typically, the models are trained with the assumption that all objects in the scene are fully observed. Previous work has shown that these models could overfit to observed agents and struggle to predict occluded agents even when they are provided to the model \citep{lange2024scene}. Furthermore, training the model to infer occluded regions actually improves the observed agents' prediction accuracy.

Recognizing the importance of joint reasoning of object trajectories and occlusions, we ask how to combine both in a single coherent framework? We propose to extend a class of structured probabilistic models called switching dynamical systems, which divides the modeling of complex continuous dynamics into a finite set/mixture of simple dynamics arbitrated by switching variables \citep{ghahramani2000variational, linderman2016recurrent}. The main attractiveness of this class of models is that it provides a unified representation where hierarchical compositions generalize to both prototype-based trajectory prediction and object-centric occlusion reasoning. For trajectory prediction, the switching variable represents the intent or behavior primitive chosen by the modeled object, where the execution of the selected intent generates trajectories prescribed by the attractor of the local dynamics. For occlusion reasoning, the switching variable represents objects' existence, which in turn modulates the prediction of their sensory measurements in combination with the scene geometry. A potential advantage of this unified yet structured framework is that it could leverage efficient inference and learning algorithms \citep{heins2024gradient} while still being amenable to manual specification of specific critical components (e.g., scene geometry). 

To demonstrate the feasibility of this framework, we evaluate a minimal implementation on the Waymo open motion dataset \citep{ettinger2021large} to predict the motion of vehicles and pedestrians in occluded traffic scenes. For trajectory prediction, we compare the model's prediction accuracy and uncertainty calibration against a few ablations. For occlusion reasoning, we visualize the model's beliefs of potential occluded pedestrian positions to show how uncertainty is maintained over time.

\section{Traffic scene modeling with switching dynamical systems}
In this section, we describe how trajectory prediction, occlusion reasoning, and object tracking can all be modeled under the unified framework of switching dynamical systems. Furthermore, each of these tasks can be modeled relatively independently, allowing us to use a divide-and-conquer approach to tackle the joint problem.

\subsection{Trajectory prediction with switching dynamical systems}
Predicting the trajectories of observed objects, or all objects when they become observed, is the most fundamental building block for any social navigation system. This setting is similar to SOTA models for fully observable predictions \citep{chai2019multipath, shi2024mtr++}. However, our focus is to show the generality of switching systems and compatibility with SOTA frameworks. 

Formally, the task of trajectory prediction is to predict the future states of $N$ objects in a scene $\mathbf{s}_{1:T}, \mathbf{s} = [s_{1}, ..., s_{N}]$ for a time horizon $T$ given past object states $\mathbf{s}_{-C:0}$ with history (or context) length $C$ and optionally auxiliary information $u_{-C:0}$, such as a high definition map as context. The state of an object typically consists of its position, velocity, and heading in the euclidean space: $[x, y, v_{x}, v_{y}, \theta]$. 

An appealing property of traffic agent trajectories found by prior work is that they tend to exhibit a spatial clustering effect where each agent's future trajectories tend to be small variations of a small number of primitive trajectories \citep{chai2019multipath}. These primitive trajectories can be interpreted as the agent's intent, such as going straight, speeding up, and turning. Thus, prior works have mostly modeled distributions over agent trajectories using Gaussian mixtures, where the mixing components correspond to the intents \citep{chai2019multipath, shi2024mtr++}. Since agent intent usually depends on the global and local context they are in, such as the movements of neighboring agents, connectivity of nearby roads and road features, and the area they are in, SOTA models typically use the Transformer architecture to condition the prediction of intent and local trajectory refinement on this auxiliary information. For simplicity, we defer conditioning on auxiliary information to future work.

The choice of modeling trajectories as Gaussian mixture densities over agent future states makes the model inherently open-loop, which has been shown to compromise local consistency in the trajectories, i.e., adjacent steps in the prediction do not follow realistic vehicle or object dynamics \citep{varadarajan2022multipath++}. The open-loop architecture is also less amenable to online filtering applications, which we consider jointly with occlusion reasoning and object tracking. To this end, we model agent trajectories using a recurrent switching linear dynamical system (rSLDS; \citealp{linderman2016recurrent}). Similar to Gaussian mixtures, the model consists of a discrete switching variable $z \in \{1, ..., |\mathcal{Z}|\}$ representing the intent of an object. However, rather than conditioning the entire predicted trajectory on a single switching variable at the beginning of the sequence, the switching variable is updated on the fly depending on the state of the object. This autoregressive updating process can be written as:
\begin{align}\label{eq:rslds}
    P(s_{1:T}) = \int_{z_{1:T}}\prod_{t=1}^{T}P(s_{t}|s_{t-1}, z_{t})P(z_{t}|z_{t-1}, s_{t-1})
\end{align}
Effectively, the sequence of switching variables can be understood as the object making dynamic discrete choices depending on the actual state it has encountered. These switching variables are not observable in filtering applications or the training dataset. Thus, we perform inference and model parameter learning using variational Bayes and variational expectation maximization \citep{beal2003variational}.

Although we do not focus on tackling the following issues in the current work, we remark that the capability of this class of models can be further enhanced, perhaps in combination with SOTA techniques, with appropriately engineered geometry and map features and multi-agent interaction modeling components to account for the spatial awareness and game-theoretic nature of driving \citep{liu2023graph, rhinehart2019precog, hallgarten2024stay, liang2020learning}. For simplicity, we consider modeling each agent independently in their own egocentric reference frame.

\subsection{Occlusion reasoning and object tracking with switching dynamical systems}
In this section, we show how occlusion reasoning and object tracking can also be modeled under the framework of switching dynamical systems.

We refer to occlusion reasoning as computing the probability of whether an object exists at a location of interest given past observations. Our model is based on \citep{engstrom2024resolving}, which considers the setting of a parked vehicle occluding the view of a potential pedestrian.

\paragraph{Single object tracking} Let the existence of a pedestrian next to the head of an occluding vehicle be denoted by a binary variable $I \in \{0, 1\}$. 
The probability that a pedestrian exists at the beginning of an episode is given by $P(I)$ and $I$ does not change over the episode, i.e., $P(I'|I)$ is identity. The pose of the ego vehicle and the pedestrian are denoted by $s^{ego}$ and $s^{ped}$, respectively. The dynamics of the ego vehicle are controlled by its own inputs $a$ according to $P(s'^{ego}|s^{ego}, a)$. The dynamics of the pedestrian follow $P(s'^{ped}|s^{ped}, I)$, where we condition it on $I$ because if the pedestrian does not exist, its state is padded with some chosen values. It is clear that the variable $I$ serves as the switching variable and can be understood as the slot for the pedestrian in slot-based object-centric models \citep{locatello2020object}.

Tracking the state of the pedestrian, including whether it exists, requires the model to interpret raw sensory observations, which depend on whether the potential pedestrian is observable given the relative geometry of the ego vehicle and the occluding object. \cite{engstrom2024resolving} model the relative geometry using a line-of-sight (LOS) method, where the pedestrian is not observable if the LOS connecting the ego vehicle and the pedestrian intersects with the occluding vehicle. Formally, this is given by the following equation where $v \in \{0, 1\}$ denotes whether the pedestrian is observable, assuming the occluding vehicle completely obscures the left-hand side of the map:
\begin{align} \label{eq:visibility_checking_fn}
    v(s^{ego}, s^{ped}, s^{obj}) = \left\{\begin{array}{cl}
        1 & \text{ if } y^{ped} - y^{obj} > \frac{y^{obj} - y^{ego}}{x^{obj} - x^{ego}}(x^{ped} - x^{obj}) \\
        0 & \text{ otherwise}
        \end{array}\right.
\end{align}

Let a binary variable $o^{I} \in \{0, 1\}$ denote whether the pedestrian is observed. Assuming the occluding object is static (and thus $x^{obj}, y^{obj}$ are constant), $o^{I}$ depends on the existence of the object, the visibility and thus both $s^{ego}$ and $s^{ped}$, i.e., $P(o^{I}|I, s^{ego}, s^{ped})$. The dependency is in fact deterministic and can be computed from the following rule:
\begin{align}\label{eq:los_geometry}
\begin{split}
    o^{I} = \left\{\begin{array}{cl}
        1 & \text{ if } I = 1 \text{ and } v(s^{ego}, s^{ped}, s^{obj}) = 1 \\
        0 & \text{ otherwise}
    \end{array}\right.
\end{split}
\end{align}
A similar mechanism is used to model observation distribution over the continuous pedestrian state $P(o^{ped}|s^{ped}, s^{ego}, I)$. where the observed pedestrian state values are equal to the padding values before the LOS and the actual state values after the LOS. The observation of ego state is given by distribution $P(o^{ego}|s^{ego})$, which, assuming negligible noise, is simply equal to the true ego state. All distributions in the model are summarized below:
\begin{align}\label{eq:los_factorization}
\begin{split}
    &P(I’|I) \\
    &P(s’^{ego}|s^{ego}, a^{ego}) \\
    &P(s’^{ped}|s^{ped}, I)\\
\end{split}
\begin{split}
    &P(o^{I}|I, s^{ego}, s^{ped}) \\
    &P(o^{ego}|s^{ego}) \\
    &P(o^{ped}|s^{ped}, s^{ego}, I)\\
\end{split}
\end{align}
Notice that this model departs slightly from regular SLDS and rSLDS in that the observation space consists of mixed discrete-continuous variables. However, the required modification is minor. 

\paragraph{Multi object tracking} For general traffic scenes, we do not know a priori how many pedestrians, if any at all, are potentially occluded by larger objects such as static and moving vehicles. We thus propose an extension of the model presented by \cite{engstrom2024resolving}, where we maintain $M$ slots for potentially occluded pedestrians, e.g., a fixed number for each region occluded by a nearby vehicle, at the beginning of each episode. 

Formally, let the states of these pedestrians be denoted by $\mathbf{s}^{ped} = \{s^{ped}_{1}, ..., s^{ped}_{M}\}$ and their existence as $\mathbf{I}= \{I_{1}, ..., I_{M}\}$. We assume, as before, the motion of these pedestrians is independent $P(s'^{ped}_{m}|s^{ped}_{m}, I_{m})$. 

The main challenge of this setting is that once a pedestrian becomes observed, either due to it stepping out of an occlusion or the ego vehicle driving past it, it is ambiguous how to associate it with and update the slots. This is because multiple slots might explain the observed pedestrian equally well, however, the pedestrian can only be associated with one of the slots. Furthermore, at any given time step, a variable number of pedestrians could be observed while others are not present or still occluded. This creates a variable-sized observation space, which is challenging to model. 

Inspired by probabilistic data association (PDA) filters \citep{rezatofighi2015joint, bar-shalom2009pda, bowman2017, wakayama2023probabilistic}, we propose an observation model with also a slot-like representation. 
Given we maintain $M$ slots for potential pedestrians, the number of observed pedestrians can be no more than $M$. Let us thus maintain $L=M$ slots for pedestrian state observations denoted with $\mathbf{o}^{ped} = \{o^{ped}_{1}, ..., o^{ped}_{L}\}$ and the discrete observation flags as $\mathbf{o}^{I} = \{o^{I}_{1}, ..., o^{I}_{L}\}$. For any $l \leq |L|$ that is not observed (e.g., they do not exist or they are still occluded), we assign the corresponding $o^{ped}_{l}$ with some padding values. For the remaining slots that are observed, the key characteristic is we do not know the correspondence between the observation slots and the occluded pedestrian slots because the observations are, in principle, permutation invariant. 

Following PDA, we assume each observation is generated by one of the pedestrian slots chosen with a uniform random probability. We denote this choice or data association variables with $c_{l} \in \{0, 1\}$. Let $\{s^{obj}_{1}, ..., s^{obj}_{N}\}$ denote the observed objects that occlude the pedestrians, we write the joint observation distribution as the following product of mixture models:
\begin{align}
\begin{split}
    P(o^{ped}_{1:L}, o^{I}_{1:L}|s^{ped}_{1:M}, I_{1:M}, s^{ego}, s^{obj}_{1:N}) &= \prod_{l=1}^{L}P(o^{ped}_{l}, o^{I}_{l}|s^{ped}_{1:M}, I_{1:M}, s^{ego}, s^{obj}_{1:N}) \\
    &= \prod_{l=1}^{L}\sum_{c^{l}}P(o^{ped}_{l}, o^{I}_{l}|c^{l}, s^{ped}_{1:M}, I_{1:M}, s^{ego}, s^{obj}_{1:N})P(c^{l}) \\
\end{split}
\end{align}

Similar to the single pedestrian setting, whether the observation values are the actual pedestrian states or the padding value depends on the visibility implied by the relative geometry. We propose a different visibility-checking method by classifying whether the pedestrian state chosen by the data association variable lies inside any occlusion polygons formed by the ego and other vehicles. We construct each polygon by connecting the LOS between the ego vehicle and each other vehicle and extending the LOS to a fixed distance (20m). The occlusion polygon is thus given by the bounding box of the other vehicle and the far points on the LOS. Arranging the line segments on the polygon in counterclockwise order, we can determine whether a pedestrian lies inside the polygon by checking whether its coordinate lies on the left-hand side of every line segment. 

At the beginning of each episode, we set the prior over pedestrian positions to be multivariate Gaussian distributions over their corresponding occlusion polygons. The pedestrian motions $P(s'^{ped}|s^{ped}, I)$ can, in principle, be modeled using rSLDS. For simplicity, we use a constant position model so that the desired inference behavior would update the Gaussians to lie inside the intersections of the occlusion polygons at adjacent time steps until pedestrians are observed or determined not present.

\subsection{Complete model}
We now put together all the building blocks introduced above into the complete model. Let us denote the closest $N$ objects at $t=0$ with states $\{s^{obj}_{1}, ..., s^{obj}_{N}\}$. These objects need no binary existence variables, and we assume they will stay fully observed throughout the prediction horizon. As before, the states of the $M$ occluded pedestrians are denoted as $\{s^{ped}_{1}, ..., s^{ped}_{M}\}$ with existence variables $\{I_{1}, ..., I_{M}\}$ and priors on their initial states and existence specified by their initial occluded regions. For the closest $N$ objects, we receive observations of their poses $\{o^{obj}_{1}, ..., o^{obj}_{N}\}$, which are identity or noisy identity mappings from the true states. For the occluded pedestrians, we observe $\{o^{ped}_{1}, ..., o^{ped}_{L}\}$ and $\{o^{I}_{1}, ..., o^{I}_{L}\}$ where a random number of them may be filled with padding values depending on the actual number of occluded objects at the current time step. 

The joint distribution of all variables (given ego actions) can be written as:
\begin{align}
\begin{split}
    &P(s^{ego}_{0:T}, \mathbf{s}^{obj}_{0:T}, \mathbf{s}^{ped}_{0:T}, \mathbf{I}_{0:T}, \mathbf{o}^{obj}_{0:T}, \mathbf{o}^{ped}_{0:T}, \mathbf{o}^{I}_{0:T}|a_{1:T}) \\
    =& \prod_{t=0}^{T}\left[P(s^{ego}_{t}|s^{ego}_{t-1}, a_{t-1}) \prod_{n=1}^{N}P(s^{obj}_{n,t}|s^{obj}_{n,t-1}) \prod_{m=1}^{M}\Big(P(s^{ped}_{m,t}|s^{ped}_{m,t-1}, I_{m, t})P(I_{m, t}|I_{m, t-1})\Big)\right] \\
    &\prod_{t=0}^{T}\left[P(o^{ego}_{t}|s^{ego}_{t}) \prod_{n=1}^{N}P(o^{obj}_{n,t}|s^{obj}_{n,t}) \prod_{l=1}^{L}\Big(P(o^{ped}_{l,t}|s^{ped}_{1:M,t}, I_{1:M, t}, s^{ego}_{t})P(o^{I}_{l, t}|I_{1:M, t}, s^{ped}_{1:M, t}, s^{ego}_{t})\Big)\right] \\
\end{split}
\end{align}

\subsection{Inference}
We use variational inference \citep{blei2017variational} to update the distribution over the existence of occluded pedestrians as well as their states which are modeled as multivariate Gaussian distributions. The full posterior we wish to obtain is $Q(s^{1:M}, I^{1:M}, c^{1:L})$. Intuitively, all pedestrian-related variables should depend on whether the pedestrian exists. We thus design the following structured posterior:
\begin{align}
    Q(s_{1:M}, I_{1:M}, c_{1:L}) = \prod_{m=1}^{M}Q(I_{m})Q(s_{m}|I_{m})\prod_{l=1}^{L}Q(c_{l})
\end{align}
When the agent does not exist, $Q(s^{m}|I^{m}=0)$ is the same as the prior which is a dirac delta distribution on the padding value and thus need not be inferred. In practice, we approximate the delta distribution using a Gaussian with small variance. 

We optimize the approximate posterior $Q$ by minimizing the variational lower bound:
\begin{align}
\begin{split}
    L(Q) &= \mathbb{E}_{Q(s_{1:M}, I_{1:M}, c_{1:L})}[\log P(o^{pose}_{1:L}, o^{pad}_{1:L}, s_{1:M}, I_{1:M}, c_{1:L}|s^{ego}, \mathbf{s}^{obj})] - \mathbb{H}[Q]\\
    &= \mathbb{E}_{Q}\left[\sum_{l=1}^{L}\big(\log P(o^{pose}_{l}, o^{pad}_{l}|c^{l}, s_{1:M}, s^{ego}, \mathbf{s}^{obj}) + \log P(c_{l})\big) + \sum_{m=1}^{M}\big(\log P(s_{m}|I_{m}) + \log P(I_{m})\big)\right] + \mathbb{H}[Q]
\end{split}
\end{align}
where $\mathbb{H}$ denotes entropy. 

The optimal posteriors have the following well-known forms \citep{blei2017variational}:
\begin{align}\label{eq:pda_posterior}
\begin{split}
    Q^{*}(I^{m}) &\propto \exp\left(\mathbb{E}_{Q^{*}(s_{m})}[\log P(s_{m}|I_{m})] + \log P(I_{m})\right) \\
    Q^{*}(s_{m}|I_{m}) &\propto \exp\left(\sum_{l=1}^{L}\mathbb{E}_{Q^{*}(s_{-m}, c_{1:L})}[\log P(o^{pose}_{l}, o^{pad}_{l}|c^{l}, s_{1:M}, s^{ego}, \mathbf{s}^{obj})] + \mathbb{E}_{Q(I_{m})}[\log P(s_{m}|I_{m})]\right) \\
    Q^{*}(c_{l}) &\propto \exp\left(\sum_{l=1}^{L}\mathbb{E}_{Q^{*}(s_{1:M}, c_{-l})}[\log P(o^{pose}_{l}, o^{pad}_{l}|c^{1:L}, s_{1:M}, s^{ego}, \mathbf{s}^{obj})] + \log P(c_{l})\right)
\end{split}
\end{align}

\section{Experiments}
We present an initial implementation of our models using the Waymo open motion dataset (WOMD; \citealp{ettinger2021large}) collected by Waymo's autonomous vehicles. The goal of the experiments is to illustrate the behavior of our models. We evaluate the trajectory prediction capability of our model quantitatively following standard practice and illustrate our model's occlusion reasoning capability using qualitative visualizations.

\paragraph{Data processing} We use the first record from the WOMD, which has 492 scenes and a total of 31319 trajectories. Each scene consists of a variable number of road users, including vehicles, pedestrians, and cyclists, for a variable time duration with a maximum trajectory length of 9.1 seconds (91 time steps). For vehicle data, we only use ego trajectories for training because there is no missing data. For pedestrian data, we filter out trajectories that lasted less than 9.1 s. This left us with 492 vehicle trajectories and 370 pedestrian trajectories in total. We use 300 vehicle trajectories and 259 pedestrian trajectories (0.7 train-test ratio) for training and the rest for evaluation. We use agent positions and velocities in the egocentric frames defined by their positions and headings at $t=0$ as features to predict. 

\paragraph{Baseline models and ablations} Since we do not model auxiliary inputs such as road graphs, we use two stripped-down version of SOTA models as our baseline: 1) a basic Gaussian mixture model (GMM) and 2) a conditional Gaussian mixture model (cGMM) which uses agent pose at the first time step to predict the parameters (i.e., means, covariance, and mixing weights) of the GMM. We use two variants of rSLDS: the full version described in (\ref{eq:rslds}) and a "recurrent-only (ro)" version where the recurrent connection does not depend on the previous switching state, i.e., $P(z'|s)$. For both GMM and rSLDS, we use a discrete state of size 10 since prior work has shown that a small number of switching variables is sufficient to capture traffic agent behavior \citep{chai2019multipath, shi2024mtr++}.

Since agent trajectories are inherently stochastic, where different turning decisions could lead to significantly different trajectories, we compare the results of providing the models with different past trajectory lengths (i.e., context) as an ablation. Longer context should lead to more accurate predictions since some of the agent intent has been realized. We consider context lengths of 10 steps and 50 steps. 

\paragraph{Evaluation metrics} We follow standard practice and measure the accuracy of trajectory predictions using average displacement error (ADE; \citealp{chai2019multipath}), defined as the mean absolute error of predicted future positions averaged over all time steps and trajectories. We are also interested in the quality of the uncertainty measures provided by the models. To this end, we use a variation of the normalized calibration error (NCE; \citealp{cao2024cctr, levi2022evaluating}), which is defined as the difference between the mean absolute error of predicted future positions and the predicted standard error averaged over all time steps and trajectories. Intuitively, this metric represents to what extent the predicted errors match with ground truth errors. 

We further adapt the metrics to account for the fact that all of the models compared generate potentially multimodal predictions. Specifically, we represent the predictive distributions of all models as Gaussian mixtures with mixing weights, means, and diagonal covariances for each future time step. These parameters are already available for the GMM prediction models. For rSLDS, we obtain the GMM predictive distribution by sampling 300 trajectories and fitting a 10-component GMM to the samples. We then compute the metrics using each GMM component and average with component metrics by the mixing weights. 

\subsection{Trajectory prediction results}
\begin{table}[!t]
    \centering
    \caption{Trajectory Prediction Results for vehicles and pedestrians measured by average displacement error (ADE) and normalized calibration error (NCE).}
    \label{tab:traj_pred}
    \renewcommand{\arraystretch}{1.25}
    \begin{tabular}{c|cccccc}
    \hline
    Model & GMM & cGMM & rSLDS-ro & rSLDS & rSLDS-c10 & rSLDS-c50 \\  
    \hline
    \multicolumn{7}{c}{Vehicle Trajectory Prediction} \\ 
    \hline
    ADE ($\downarrow$) & 13.54 & 6.76 & 4.62 & 3.75 & 3.00 & 1.02 \\
    NCE ($\downarrow$) & 11.34 & 4.74 & 3.84 & 2.67 & 2.28 & 0.82 \\ 
    \hline 
    \multicolumn{7}{c}{Pedestrian Trajectory Prediction} \\ 
    \hline
    ADE ($\downarrow$) & 1.89 & 0.95 & 1.41 & 1.57 & 1.06 & 0.42 \\
    NCE ($\downarrow$) & 1.57 & 0.97 & 1.46 & 1.19 & 0.86 & 0.34 \\
    \hline
    \end{tabular}
\end{table}

\begin{figure}[!t]
    \centering
    \includegraphics[width=0.9\linewidth]{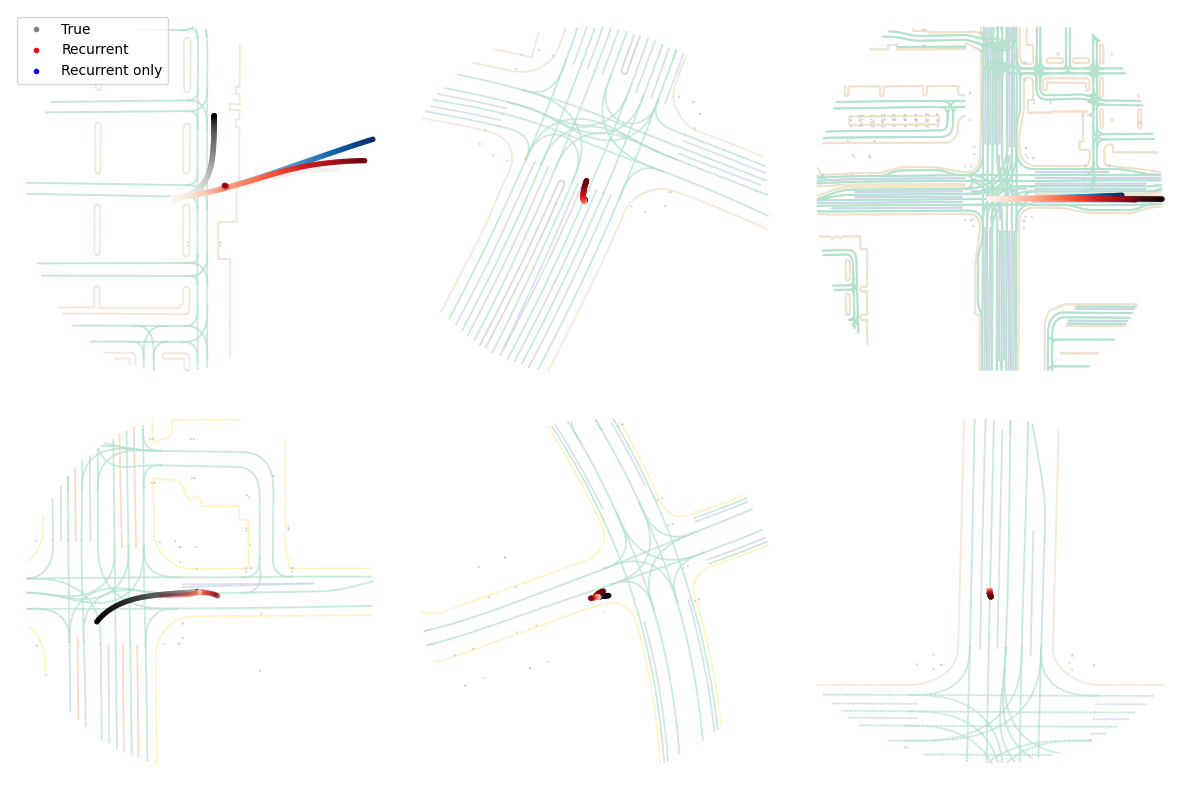}
    \caption{Visualizations of predicted vehicle trajectories.}
    \label{fig:rslds_car}
\end{figure}

\begin{figure}[!t]
    \centering
    \includegraphics[width=0.9\linewidth]{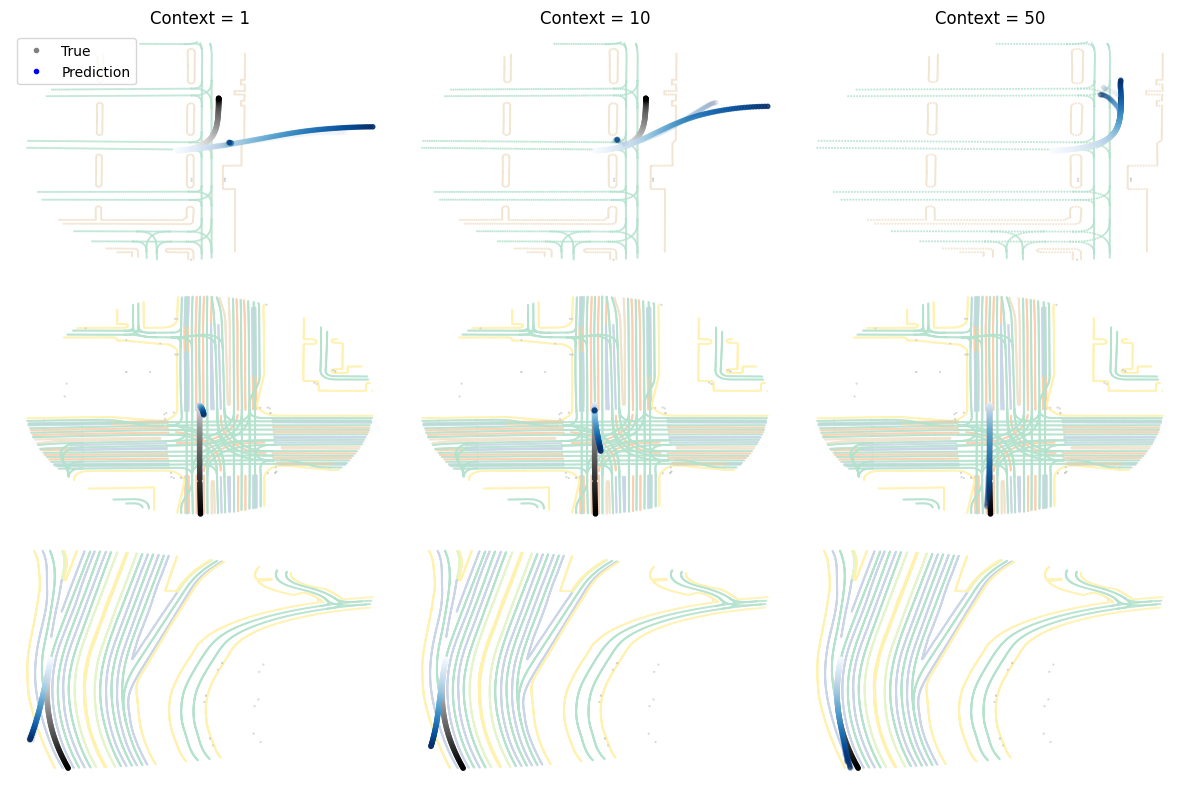}
    \caption{Visualizations of predicted vehicle trajectories by context length.}
    \label{fig:rslds_car_context}
\end{figure}

Table \ref{tab:traj_pred} shows the trajectory prediction results. rSLDS models outperformed GMMs in prediction accuracy (in terms of ADE) and uncertainty calibration (in terms of NCE) in all but one cases. For vehicle trajectory prediction, GMM performed significantly worse than other models, most likely because it doesn't consider the agent's current state or more past context. The conditional GMM -- our stripped-down version of SOTA models which conditions the prediction of GMM parameters on agent state information -- significantly improves over the basic GMM. However, its accuracy is still lower than rSLDS by 2.1 m. For pedestrian trajectory prediction, however, the difference among models are significantly smaller. In fact, cGMM achieved the highest accuracy and calibration. This is likely because most pedestrians tend to be stationary or moving at slow speed in the dataset, and when they are moving,  a constant velocity model is good enough to model their behavior. The rSLDS model may be biased towards pedestrians moving. This can be fixed in future work by training on more data or manually specifying certain model components.

A visualization of the rSLDS prediction conditioning on only the first time step is shown in Fig. \ref{fig:rslds_car}. The figure shows that the model can accurately predict vehicles traveling straight, executing small movements, or close to stationary. However, its predictions can deviate significantly from ground truth for turning trajectories. This was expected given that we do not use road features as input to the model and highlight the importance of road features in future iterations of the model. 

The lack of long-horizon prediction accuracy can, however, be compensated by the ability to frequently recompute predictions online. Fig. \ref{fig:rslds_car_context} shows three predicted vehicle trajectories when contexts of different lengths are provided. We see that more context improves prediction accuracy. This implies that for tracking and filtering applications, one might be able to trade off model complexity and accuracy vs. speed.

\subsection{Occlusion reasoning results}
\begin{figure*}[t!]
    \centering
    \begin{subfigure}[t]{0.33\textwidth}
        \centering
        \includegraphics[width=1.\textwidth]{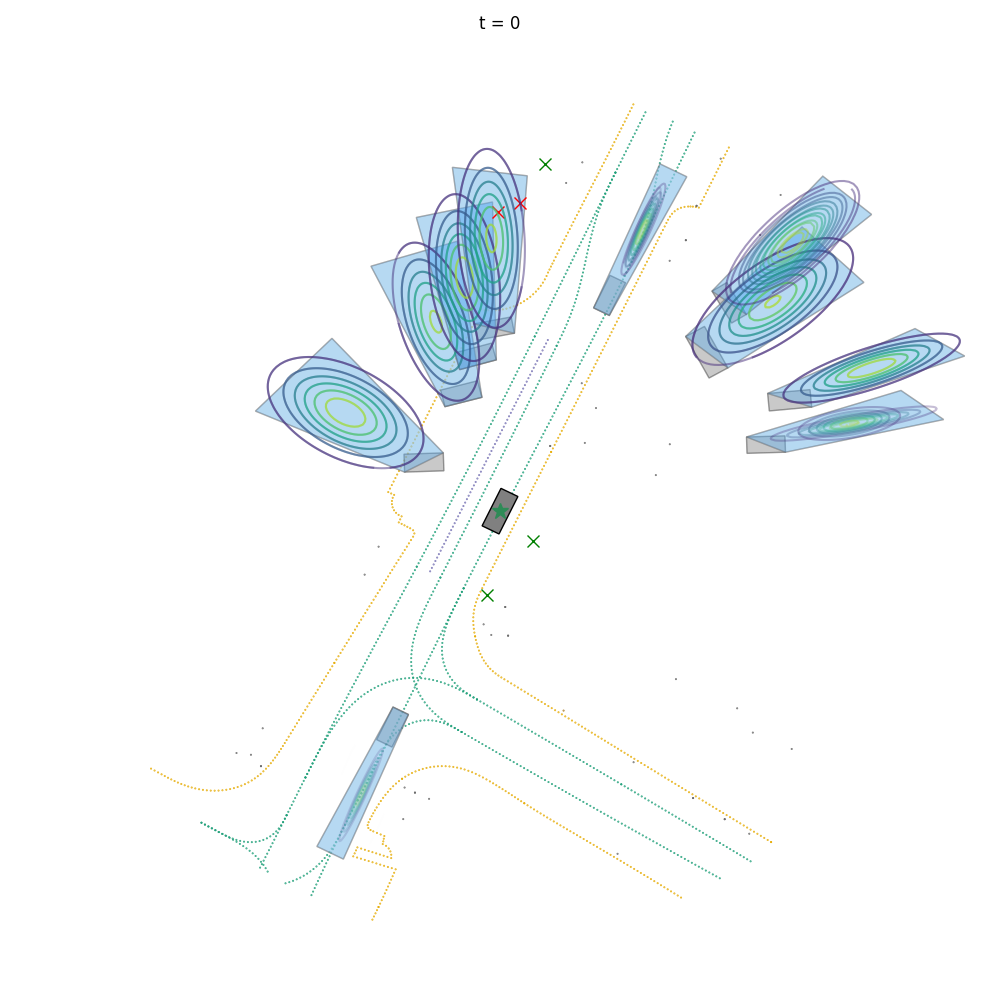}
    \end{subfigure}%
    ~ 
    \begin{subfigure}[t]{0.33\textwidth}
        \centering
        \includegraphics[width=1.\textwidth]{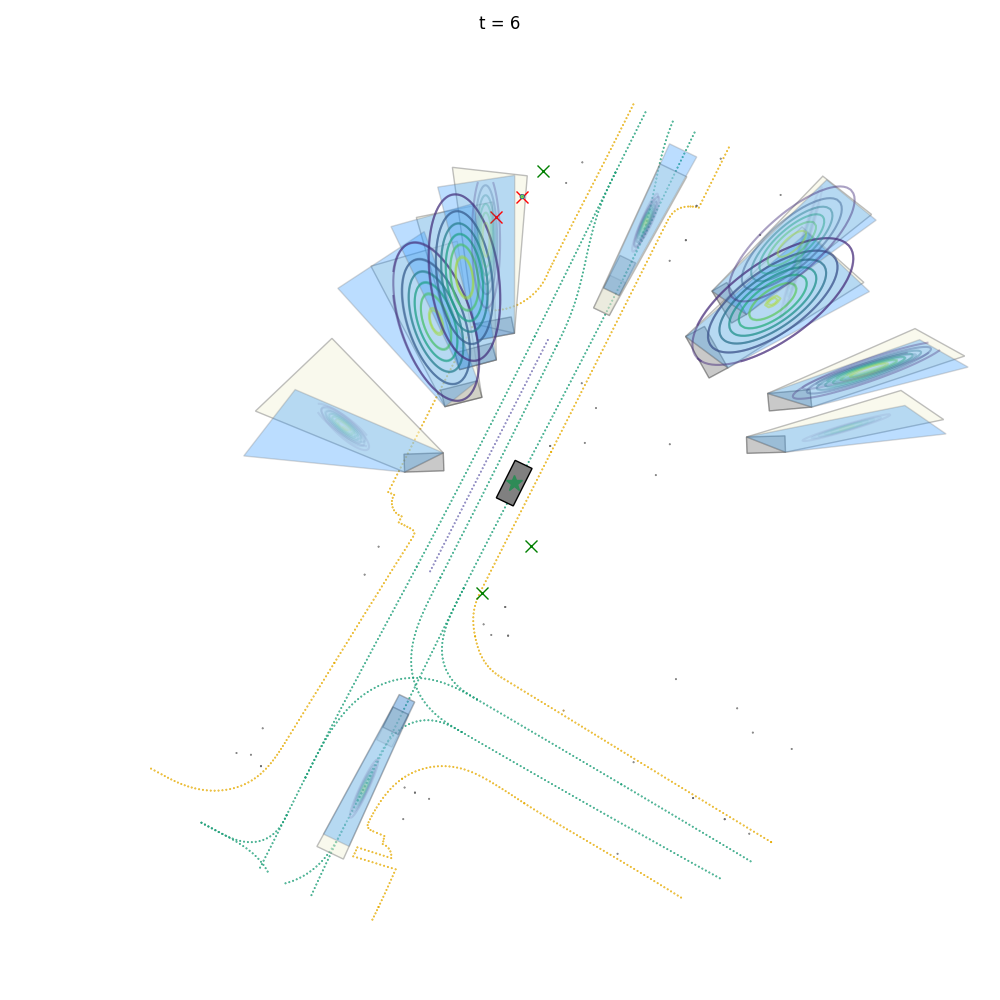}
    \end{subfigure}%
    ~
    \begin{subfigure}[t]{0.33\textwidth}
        \centering
        \includegraphics[width=1.\textwidth]{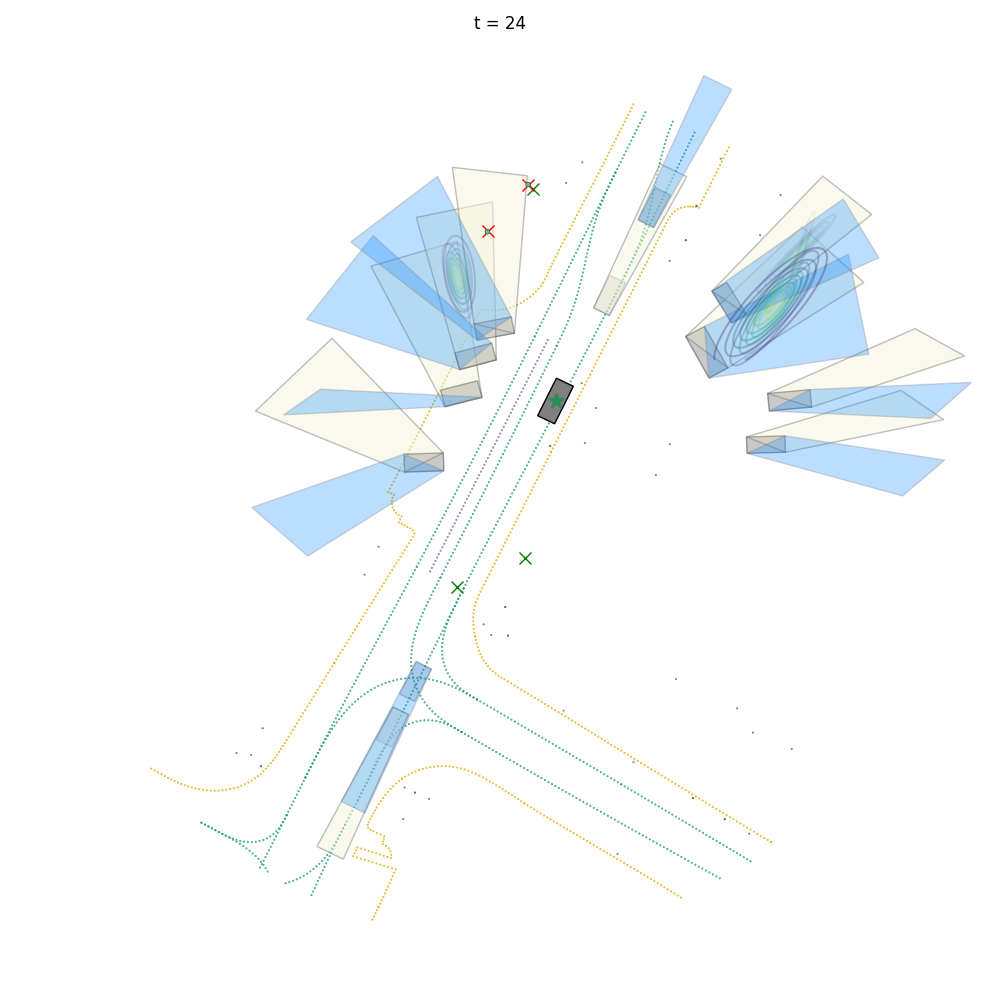}
    \end{subfigure}
    \caption{Visualization of occlusion inference. The beige-colored polygons in all plots represent the occluded regions at $t=0$. The blue-colored polygons represent the occluded regions at the time step of the respective plot. Vehicles are represented by gray boxes, and the ego vehicle is highlighted by a green star. Pedestrians are represented by crosses where observed ones at $t=0$ are colored in green and occluded ones are colored in red. The circular contours represent the Gaussian belief over pedestrian positions, and their transparency represents the model's beliefs over the existence of the corresponding pedestrian.}
    \label{fig:pda}
\end{figure*}

In this section, we demonstrate some qualitative results on the occlusion reasoning model. To show that the model can maintain uncertainty about the existence of pedestrians behind occlusions and adapt its beliefs dynamically, we selected a few representative scenarios from WOMD, where pedestrians were initially hidden behind parked or moving vehicles, which are then revealed due to the movement of either the ego or other vehicles. We assume each occluded region can hide a maximum of 2 pedestrians at $t=0$. Intuitively, if no pedestrian is observed near an occluded region, then the model's updated belief about the pedestrian positions should be the intersection of the occlusion polygons at the current and the previous time step. We use these polygon intersections to initialize the variational posteriors in (\ref{eq:pda_posterior}).

Fig. \ref{fig:pda} shows an example of the model's belief updating process. At $t=0$ (left plot), there is a set of Gaussian distributions on top of every occlusion polygon representing the model's beliefs about pedestrian positions. At $t = 6$ (middle plot), as the ego vehicle drives past several parked vehicles, a number of the Gaussians shrink in size while a subset of them have disappeared as the model believes that no pedestrian exists in that region. One of the two pedestrians occluded at $t=0$ at the top also became observable. Thus, one of the Gaussians had clamped onto their true position with small variance, representing the model's precise belief of their existence and positions. At $t=24$ (right plot), as the ego vehicle has driven past a number of occlusions, most of the Gaussians have disappeared, representing that the model believes (correctly) that no pedestrians exist behind those occlusions.

\section{Conclusion and Future Work}
In this work, we made some initial attempts towards developing structured probabilistic models for trajectory pre-
diction and occlusion reasoning. We showed that both tasks can be embedded in the same framework and yet still amenable to divide-and-conquer solutions. Our experimental results show that, when conditioned on the same information, the closed-loop rSLDS models achieved higher predictive accuracy and uncertainty calibration. In future work, we plan to incorporate auxiliary information such as road graphs to
improve prediction accuracy and implement efficient inference algorithms that are particularly suitable to this family of models \citep{heins2024gradient}.

\bibliography{ref.bib}

\begin{thebibliography}{25}
\providecommand{\natexlab}[1]{#1}
\providecommand{\url}[1]{\texttt{#1}}
\expandafter\ifx\csname urlstyle\endcsname\relax
  \providecommand{\doi}[1]{doi: #1}\else
  \providecommand{\doi}{doi: \begingroup \urlstyle{rm}\Url}\fi

\bibitem[Bar-Shalom et~al.(2009)Bar-Shalom, Daum, and Huang]{bar-shalom2009pda}
Y.~Bar-Shalom, F.~Daum, and J.~Huang.
\newblock The probabilistic data association filter.
\newblock \emph{IEEE Control Systems Magazine}, 29\penalty0 (6):\penalty0 82--100, 2009.
\newblock \doi{10.1109/MCS.2009.934469}.

\bibitem[Beal(2003)]{beal2003variational}
M.~J. Beal.
\newblock \emph{Variational algorithms for approximate Bayesian inference}.
\newblock University of London, University College London (United Kingdom), 2003.

\bibitem[Blei et~al.(2017)Blei, Kucukelbir, and McAuliffe]{blei2017variational}
D.~M. Blei, A.~Kucukelbir, and J.~D. McAuliffe.
\newblock Variational inference: A review for statisticians.
\newblock \emph{Journal of the American statistical Association}, 112\penalty0 (518):\penalty0 859--877, 2017.

\bibitem[Bowman et~al.(2017)Bowman, Atanasov, Daniilidis, and Pappas]{bowman2017}
S.~L. Bowman, N.~Atanasov, K.~Daniilidis, and G.~J. Pappas.
\newblock Probabilistic data association for semantic slam.
\newblock In \emph{2017 IEEE International Conference on Robotics and Automation (ICRA)}, pages 1722--1729, 2017.
\newblock \doi{10.1109/ICRA.2017.7989203}.

\bibitem[Cao et~al.(2024)Cao, Chen, Wang, Song, Tan, and Li]{cao2024cctr}
C.~Cao, X.~Chen, J.~Wang, Q.~Song, R.~Tan, and Y.-H. Li.
\newblock Cctr: Calibrating trajectory prediction for uncertainty-aware motion planning in autonomous driving.
\newblock In \emph{Proceedings of the AAAI Conference on Artificial Intelligence}, volume~38, pages 20949--20957, 2024.

\bibitem[Chai et~al.(2019)Chai, Sapp, Bansal, and Anguelov]{chai2019multipath}
Y.~Chai, B.~Sapp, M.~Bansal, and D.~Anguelov.
\newblock Multipath: Multiple probabilistic anchor trajectory hypotheses for behavior prediction.
\newblock \emph{arXiv preprint arXiv:1910.05449}, 2019.

\bibitem[Engstr{\"o}m et~al.(2024)Engstr{\"o}m, Wei, McDonald, Garcia, O'Kelly, and Johnson]{engstrom2024resolving}
J.~Engstr{\"o}m, R.~Wei, A.~D. McDonald, A.~Garcia, M.~O'Kelly, and L.~Johnson.
\newblock Resolving uncertainty on the fly: modeling adaptive driving behavior as active inference.
\newblock \emph{Frontiers in neurorobotics}, 18:\penalty0 1341750, 2024.

\bibitem[Ettinger et~al.(2021)Ettinger, Cheng, Caine, Liu, Zhao, Pradhan, Chai, Sapp, Qi, Zhou, et~al.]{ettinger2021large}
S.~Ettinger, S.~Cheng, B.~Caine, C.~Liu, H.~Zhao, S.~Pradhan, Y.~Chai, B.~Sapp, C.~R. Qi, Y.~Zhou, et~al.
\newblock Large scale interactive motion forecasting for autonomous driving: The waymo open motion dataset.
\newblock In \emph{Proceedings of the IEEE/CVF International Conference on Computer Vision}, pages 9710--9719, 2021.

\bibitem[Ghahramani and Hinton(2000)]{ghahramani2000variational}
Z.~Ghahramani and G.~E. Hinton.
\newblock Variational learning for switching state-space models.
\newblock \emph{Neural computation}, 12\penalty0 (4):\penalty0 831--864, 2000.

\bibitem[Hallgarten et~al.(2024)Hallgarten, Kisa, Stoll, and Zell]{hallgarten2024stay}
M.~Hallgarten, I.~Kisa, M.~Stoll, and A.~Zell.
\newblock Stay on track: A frenet wrapper to overcome off-road trajectories in vehicle motion prediction.
\newblock In \emph{2024 IEEE Intelligent Vehicles Symposium (IV)}, pages 795--802. IEEE, 2024.

\bibitem[Heins et~al.(2024)Heins, Wu, Markovic, Tschantz, Beck, and Buckley]{heins2024gradient}
C.~Heins, H.~Wu, D.~Markovic, A.~Tschantz, J.~Beck, and C.~Buckley.
\newblock Gradient-free variational learning with conditional mixture networks.
\newblock \emph{arXiv preprint arXiv:2408.16429}, 2024.

\bibitem[Lange et~al.(2024)Lange, Li, and Kochenderfer]{lange2024scene}
B.~Lange, J.~Li, and M.~J. Kochenderfer.
\newblock Scene informer: Anchor-based occlusion inference and trajectory prediction in partially observable environments.
\newblock In \emph{2024 IEEE International Conference on Robotics and Automation (ICRA)}, pages 14138--14145. IEEE, 2024.

\bibitem[Levi et~al.(2022)Levi, Gispan, Giladi, and Fetaya]{levi2022evaluating}
D.~Levi, L.~Gispan, N.~Giladi, and E.~Fetaya.
\newblock Evaluating and calibrating uncertainty prediction in regression tasks.
\newblock \emph{Sensors}, 22\penalty0 (15):\penalty0 5540, 2022.

\bibitem[Liang et~al.(2020)Liang, Yang, Hu, Chen, Liao, Feng, and Urtasun]{liang2020learning}
M.~Liang, B.~Yang, R.~Hu, Y.~Chen, R.~Liao, S.~Feng, and R.~Urtasun.
\newblock Learning lane graph representations for motion forecasting.
\newblock In \emph{Computer Vision--ECCV 2020: 16th European Conference, Glasgow, UK, August 23--28, 2020, Proceedings, Part II 16}, pages 541--556. Springer, 2020.

\bibitem[Linderman et~al.(2016)Linderman, Miller, Adams, Blei, Paninski, and Johnson]{linderman2016recurrent}
S.~W. Linderman, A.~C. Miller, R.~P. Adams, D.~M. Blei, L.~Paninski, and M.~J. Johnson.
\newblock Recurrent switching linear dynamical systems.
\newblock \emph{arXiv preprint arXiv:1610.08466}, 2016.

\bibitem[Liu et~al.(2023)Liu, Magliacane, Kofinas, and Gavves]{liu2023graph}
Y.~Liu, S.~Magliacane, M.~Kofinas, and E.~Gavves.
\newblock Graph switching dynamical systems.
\newblock In \emph{International Conference on Machine Learning}, pages 21867--21883. PMLR, 2023.

\bibitem[Locatello et~al.(2020)Locatello, Weissenborn, Unterthiner, Mahendran, Heigold, Uszkoreit, Dosovitskiy, and Kipf]{locatello2020object}
F.~Locatello, D.~Weissenborn, T.~Unterthiner, A.~Mahendran, G.~Heigold, J.~Uszkoreit, A.~Dosovitskiy, and T.~Kipf.
\newblock Object-centric learning with slot attention.
\newblock \emph{Advances in neural information processing systems}, 33:\penalty0 11525--11538, 2020.

\bibitem[Nayakanti et~al.(2023)Nayakanti, Al-Rfou, Zhou, Goel, Refaat, and Sapp]{nayakanti2023wayformer}
N.~Nayakanti, R.~Al-Rfou, A.~Zhou, K.~Goel, K.~S. Refaat, and B.~Sapp.
\newblock Wayformer: Motion forecasting via simple \& efficient attention networks.
\newblock In \emph{2023 IEEE International Conference on Robotics and Automation (ICRA)}, pages 2980--2987. IEEE, 2023.

\bibitem[Pedregosa et~al.(2011)Pedregosa, Varoquaux, Gramfort, Michel, Thirion, Grisel, Blondel, Prettenhofer, Weiss, Dubourg, Vanderplas, Passos, Cournapeau, Brucher, Perrot, and Duchesnay]{scikit-learn}
F.~Pedregosa, G.~Varoquaux, A.~Gramfort, V.~Michel, B.~Thirion, O.~Grisel, M.~Blondel, P.~Prettenhofer, R.~Weiss, V.~Dubourg, J.~Vanderplas, A.~Passos, D.~Cournapeau, M.~Brucher, M.~Perrot, and E.~Duchesnay.
\newblock Scikit-learn: Machine learning in {P}ython.
\newblock \emph{Journal of Machine Learning Research}, 12:\penalty0 2825--2830, 2011.

\bibitem[Rezatofighi et~al.(2015)Rezatofighi, Milan, Zhang, Shi, Dick, and Reid]{rezatofighi2015joint}
S.~H. Rezatofighi, A.~Milan, Z.~Zhang, Q.~Shi, A.~Dick, and I.~Reid.
\newblock Joint probabilistic data association revisited.
\newblock In \emph{Proceedings of the IEEE international conference on computer vision}, pages 3047--3055, 2015.

\bibitem[Rhinehart et~al.(2019)Rhinehart, McAllister, Kitani, and Levine]{rhinehart2019precog}
N.~Rhinehart, R.~McAllister, K.~Kitani, and S.~Levine.
\newblock Precog: Prediction conditioned on goals in visual multi-agent settings.
\newblock In \emph{Proceedings of the IEEE/CVF International Conference on Computer Vision}, pages 2821--2830, 2019.

\bibitem[Shi et~al.(2024)Shi, Jiang, Dai, and Schiele]{shi2024mtr++}
S.~Shi, L.~Jiang, D.~Dai, and B.~Schiele.
\newblock Mtr++: Multi-agent motion prediction with symmetric scene modeling and guided intention querying.
\newblock \emph{IEEE Transactions on Pattern Analysis and Machine Intelligence}, 2024.

\bibitem[Trautman and Krause(2010)]{trautman2010unfreezing}
P.~Trautman and A.~Krause.
\newblock Unfreezing the robot: Navigation in dense, interacting crowds.
\newblock In \emph{2010 IEEE/RSJ International Conference on Intelligent Robots and Systems}, pages 797--803. IEEE, 2010.

\bibitem[Varadarajan et~al.(2022)Varadarajan, Hefny, Srivastava, Refaat, Nayakanti, Cornman, Chen, Douillard, Lam, Anguelov, et~al.]{varadarajan2022multipath++}
B.~Varadarajan, A.~Hefny, A.~Srivastava, K.~S. Refaat, N.~Nayakanti, A.~Cornman, K.~Chen, B.~Douillard, C.~P. Lam, D.~Anguelov, et~al.
\newblock Multipath++: Efficient information fusion and trajectory aggregation for behavior prediction.
\newblock In \emph{2022 International Conference on Robotics and Automation (ICRA)}, pages 7814--7821. IEEE, 2022.

\bibitem[Wakayama and Ahmed(2023)]{wakayama2023probabilistic}
S.~Wakayama and N.~Ahmed.
\newblock Probabilistic semantic data association for collaborative human-robot sensing.
\newblock \emph{IEEE Transactions on Robotics}, 39\penalty0 (4):\penalty0 3008--3023, 2023.

\end{thebibliography}

\newpage
\appendix
\section{Implementation detail}
We use the ssm~\footnote{\url{https://github.com/lindermanlab/ssm}} library to train the rSLDS models using the Laplace-EM algorithm. We train the GMMs using the Scikit-Learn library \citep{scikit-learn}. For conditional GMMs, we initialize with the trained GMM parameters and use the current agent pose to predict the residuals of the GMM's means, log standard deviations, and mixing weight logits using linear layers. For example, the predicted logits are obtained as $\texttt{logits} = \texttt{gmm\_logits} + \texttt{linear(agent\_pose)}$. We train these linear layers using full batch gradient descent. 

\section{Additional results}
\subsection{Pedestrian trajectory predictions}
\begin{figure}[!thb]
    \centering
    \includegraphics[width=0.9\linewidth]{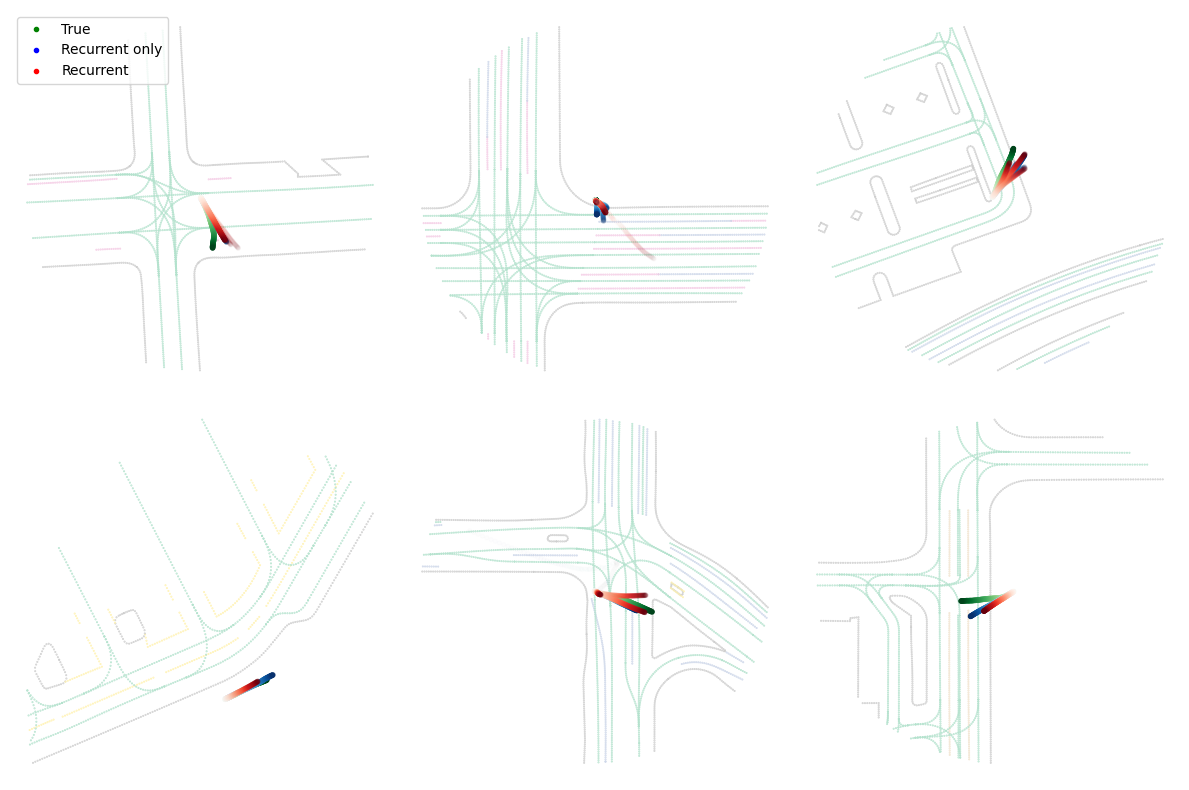}
    \caption{Visualizations of predicted pedestrian trajectories.}
    \label{fig:rslds_ped}
\end{figure}

\begin{figure}[!thb]
    \centering
    \includegraphics[width=0.9\linewidth]{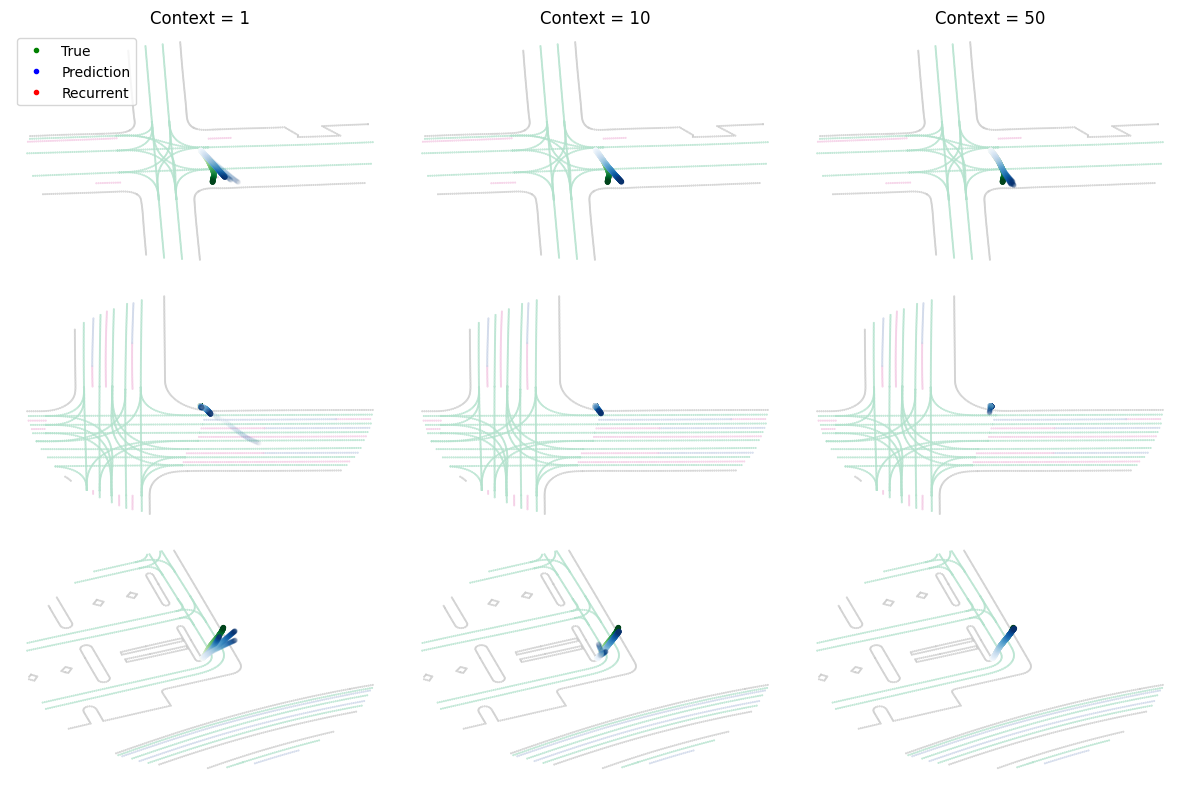}
    \caption{Visualizations of predicted pedestrian trajectories by context length.}
    \label{fig:rslds_ped_context}
\end{figure}

Visualizations of pedestrian trajectory prediction and that using different context lengths are shown in Fig. \ref{fig:rslds_ped} and \ref{fig:rslds_ped_context}, respectively.

\end{document}